# Rapid detection of soil carbonates by means of NIR spectroscopy, deep learning methods and phase quantification by powder X-ray diffraction.

Lykourgos Chiniadis[a], Petros Tamvakis[a]

[a] Athena Research and Innovation Centre, ILSP Xanthi's Division, Xanthi 67100, Greece,

e-mail: Lykourgos Chiniadis: lykchiniadis@gmail.com and Petros Tamvakis: petamv77@gmail.com

ORCID(s): LC 0000-0002-7118-9942, PT 0000-0001-9514-8283

Authorship contribution statement

ABSTRACT

Soil near-Infrared (NIR) spectral absorbance/reflectance libraries are utilized towards improving agricultural production and analysis of soil properties which are key prerequisite for agro-ecological balance and environmental sustainability. Carbonates in particular, represent a soil property which is mostly affected even by mild, let alone extreme, changes of environmental conditions during climate change. In this study we propose a rapid and efficient way to predict carbonates content in soil by means of Fourier Transform Near-Infrared (FT-NIR) reflectance spectroscopy and by use of deep learning methods. We exploited multiple machine learning methods, such as: 1) a Multi-Layered Perceptron Regressor (MLP) and 2) a Convolutional Neural Network (CNN) and compare their performance with other traditional machine learning algorithms such as Partial Least Squares Regression (PLSR), Cubist and Support Vector Machines (SVM) on the combined dataset of two NIR spectral libraries: Kellogg Soil Survey Laboratory (KSSL) of the United States Department of Agriculture (USDA), a dataset of soil samples reflectance spectra collected nationwide, and Land Use and Coverage Area Frame Survey (LUCAS) TopSoil (European Soil Library) which contains soil sample absorbance spectra from all over the European Union, and use them to predict carbonate content on never-before-seen soil samples. Soil samples in KSSL and in TopSoil spectral libraries were acquired in the spectral region of visible–near infrared (Vis-NIR) (350-2500 nm), however in this study, only the NIR spectral region (1150-2500 nm) was utilized. Quantification of carbonates by means of X-ray-Diffraction is in good agreement with the volumetric method and the MLP prediction. Our work contributes to rapid carbonates content prediction in soil samples in cases where: 1) no volumetric method is available and 2) only NIR spectra absorbance data are available. Up till now and to the best of our knowledge, there exists no other study, that presents a prediction model trained on such an extensive dataset with such promising results on unseen data, undoubtedly supporting the notion that deep learning models present excellent prediction tools for soil carbonates content.

## 1. Introduction

NIR spectroscopy is a rapid, non-destructive method of low cost that provides excellent correlation of observed and predicted values when regression algorithms are applied in spectral data and especially in large datasets (Bellon-Maurel and McBratney, 2011; Du and Zhou, 2009; Guerrero et al., 2016; Shepherd and Walsh, 2007). Early research studies in soil properties are using minimal input in terms of soil samples collected and considering minimal geographical areas and locations (Ben-Dor and Banin, 1995; Bertrand et al., 2002; Bogrekci and Lee, 2005; Brown et al., 2005; Chang et al., 2001; Dalal and Henry, 1986; Islam et al., 2003). These studies typically modeled total carbon (tC) and soil organic carbon (SOC) (Askari et al., 2018; Bai et al., 2022; Du and Zhou, 2009; Fontán et al., 2011; Genot et al., 2011; Jia et al., 2014; Knadel et al., 2013; Miltz and Don, 2012; Morón and Cozzolino, 2002; Peng et al., 2013; Reeves and Van Kessel, 1999a), soil organic matter (SOM) (Wang and Pan, 2016), soil inorganic carbon and carbonates (SIC) (Barthès et al., 2016a; Khayamim et al., 2015), total nitrogen (tN) (Cho et al., 1998; Genot et al., 2011; Jia et al., 2014; Miltz and Don, 2012; Morón and Cozzolino, 2002; Reeves et al., 1999; Reeves and Van Kessel, 1999b), phosphorous content (Jia et al., 2014), potassium content (Jia et al., 2014), clay (Genot et al., 2011; Knadel et al., 2013), pH (Morón and Cozzolino, 2002; Reeves et al., 1999), moisture content (Dalal and Henry, 1986; Hummel et al., 2001; Morellos et al., 2016; Mouazen et al., 2005) and cationic exchange capacity (CEC) (Chodak et al., 2004; Genot et al., 2011). Some studies had also modeled other properties specific to their objectives, such as total elemental content e.g., (Reeves and Smith, 2009), extractable nutrients e.g., (Vašát et al., 2014), biological properties e.g., (Bellino et al., 2016). Computational progress and especially new algorithms applied in the spectral data are nowadays efficient to predict soil properties for larger datasets and fields, acquiring hundreds and thousands soil samples over locations of interest to generate large databases of spectral and physicochemical properties. Recently, hyperspectral NIR spectral data acquisition is even possible using satellites with high resolution data acquired (Gomez et al., 2008; Liu et al., 2018).

A large spectral database with spectra from all soil types (Guo et al., 2003) is the Kellogg Soil Survey Laboratory (KSSL), of Natural Resource Conservation Service (NRSC) under the United States Department of Agriculture (USDA) (https://ncsslabdatamart.sc.egov.usda.gov/), has been a service agency that is collecting spectral and other data, is measuring physicochemical soil properties and is uploading and sharing the data, since 80 and more years until nowadays. KSSL is recording data from physicochemical properties and location characteristics to vis-NIR and MIR spectra across United States of America. Other similar spectral libraries include the European LUCAS Topsoil dataset (Orgiazzi et al., 2018) (https://esdac.jrc.ec.europa.eu/content/lucas2015-topsoil-data), the Brazilian Soil Spectral Library (BSSL) (Demattê et al., 2019), global soil datasets (Viscarra Rossel et al., 2016) and the Chinese vis–NIR soil spectral library (CSSL) (Shi et al., 2015). Also, combination of spectral libraries is feasible in terms of applying specific standards and statistic protocols (Ben Dor et al., 2015).

**Related Work**

Machine Learning (ML) techniques were introduced in chemometrics and soil spectroscopy as a means for information extraction and soil properties prediction. Both traditional ML and Deep Learning (DL) approaches have been extensively adopted (Morellos et al., 2016; Stevens et al., 2013; Viscarra Rossel et al., 2016) because of their intrinsic property to handle high dimensional data such as spectra. Deep neural networks, in particular, are capable of handling data in almost every form e.g. tabular data, text, images and audio, coupled with recent increases in computational power and the expansion of system memory capabilities, now make possible the use of spectral data to perform spectrogram-based analysis for soil properties predictions. As there are usually thousands of predictor variables (reflectance values), neural networks make an excellent choice for such a task because of their intrinsic trait to handle well data that suffer from high dimensionality. Convolutional Neural Networks (CNN) (Lecun, 1989) have been used in image recognition with remarkable success and constitute one of the most powerful DL techniques for modeling complex processes such as pattern recognition in image based applications. Recently, Tsakiridis et al., (2020) utilized CNNs to predict multiple soil properties. The successful use of CNNs for soil property prediction without the use of pre-processed spectra was demonstrated in (Padarian et al., 2019). They proposed the representation of raw spectral data as a two-dimensional (2D) spectrogram and showed its superior performance over traditional ML techniques such as Partial least squares regression (PLSR) (Martens and Næs, 1989) and Cubist regression trees (Quinlan, 1992). Ng et al., (2019) used the combined spectra of soil samples to train 1D and 2D CNNs that both outperformed traditional ML techniques.

In our study, we compare different methodologies and techniques based on both computational regression techniques on NIR spectral data and experimental methods that usually apply on soil samples. Neural Networks algorithms outperformed the linear models for the soil carbonates content prediction as one term, and secondly, in the prediction of important NIR peaks. Classic mineralogical and analytical methods were also applied in soil samples establishing the feasibility of introducing neural networks regression methods in soil NIR spectra.

Much effort is given towards elucidating soil properties in all continents and all recorded data is valuable knowledge for today and future research. Moreover, data science applied to spectral and physicochemical libraries for the construction of calibration libraries is a novel approach redefining universal scientific effort. Non-destructive and scientific effective methods are a new path in modern soil research and is adding value from a spectroscopic point of view.

## 2. Methodology

### 2.1. Sample pretreatment

The diffuse NIR reflectance spectra of air-dried and (<250 μm) sieved soil samples were measured in the laboratory. Before spectral measurements, the samples were placed in glass Petri dishes and spectral acquisition was performed with thickness of the sample of ~3-4 cm to avoid transmission effects, for all samples. Background spectra was subtracted from all acquired NIR spectra.

### 2.2. Spectral acquisition

A Perkin-Elmer Spectrum N Two FT-NIR spectrophotometer (quartz beamsplitter and $LiTaO_3$ NIR detector based on diode array) was used for spectral measurements in the spectral region of 1150–2500 nm. The spectral data was screened to ensure percentage reflectance measurements did not exceed theoretical limits, from 0.0 to 100.0. All spectra were exported to wavelength intervals of 0.5 nm. Absorbance spectra were calculated by means of equation (1).

### 2.3. Pretreatment of spectral libraries as calibration dataset

The soil spectra were transformed using three pretreatment methods prior to chemometric modeling, as the best treatment was not known a priori. This technique included the (pseudo) absorbance transformation, normalization of spectra between values 0-1, Savitzky-Golay smoothing filtering with a window size of 11, polyorder of 2 for the differentiating SG-1 and window size of 13, polyorder of 2 for the differentiating SG-2 (Savitzky and Golay, 1964) were applied to the spectral data to reduce noise and enhance spectral features to all calibration data of both KSSL-LUCAS TopSoil libraries combined, in order to reduce undesirable variance of the data to improve the predictive capacity of the calibration models. These procedures were applied to the original reflectance spectra (R) for KSSL (USDA) and TopSoil (Europe) and transformed to Absorbance (A) by the following equation:

$$A = \log_{10}(1/R) \tag{1}$$

No transmission (T) is observed due to the large thickness of the sample measured with the NIR spectrophotometer. Data screening resulted 28,615 samples for modeling with PLSR, Cubist, SVM, MLP and CNN models. Spectra of KSSL-USDA spectral library were transformed from percent reflectance to absorbance, by means of equation (1). The values of $CaCO_3$ were normalized to g/100g since LUCAS-TopSoil laboratory measurement unit of $CaCO_3$ is in g/kg.

In machine learning applications, it is a well-known fact that the size of the training set plays a crucial role in generalization. Generalization is the ability of the model to perform well on never-before-seen data. Typically, we aim for a large, diverse training set and to that end a decision was made to merge the two spectral libraries into a larger one. Before merging, we compare both spectral datasets based on the Wasserstein metric which is a common way to compare the probability distributions of two variables:

$$W_p(P, Q) = \left(\frac{1}{n}\sum_{i=1}^{n} \left|\left|X_{(i)} - Y_{(i)}\right|\right|^p\right)^{\frac{1}{p}} \tag{2}$$

where *P*, *Q* are probability distributions (one-dimensional) and *p* is the number of moments. The results show that the two libraries are similar enough to be merged (Wasserstein distance=1.78). Comparison of the new combined training dataset with the test set shows low similarity between them (Wasserstein distance=6.48), which raises a concern of how well our model will perform on the test set.

**2.4. Study Area**

The investigated area falls within the so-called areas of wine fields of Vourvoukelis winery in Avdira in Xanthi, Greece (40º 54' 20'' N, 25º 48' 24''E) and of organic wine fields of Xatzisavva winery in Alexandroupolis, Greece (41º 00' 09'' N, 24º 55' 00'' E).

**2.5. Laboratory carbonates measurement**

Results throughout this study are given as $CO_3^{2-}$ equivalents. The measurement is based on measuring the gaseous $CO_2$ released from carbonate reacting with 6M HCl and is expected to be equally effective at measuring soil carbonates released from the full range of carbonate soil minerals including common forms such as calcite, magnesite, and dolomite and their hydrous compounds.

Carbonates equivalents were determined at both locations by pressure calcimeter method treating the <250 μm soil fraction with 6M HCl in a closed vial. At KSSL this was volumetric method 4E1a1a1 (https://www.nrcs.usda.gov/sites/default/files/2023-01/SSIR42.pdf, accessed on 20 November 2022), and at LUCAS-Topsoil this method was similar volumetric method with ISO 10693:1995 (https://esdac.jrc.ec.europa.eu/public_path/shared_folder/dataset/66/JRC121325_lucas_2015_topsoil_survey_final_1.pdf, accessed on 21 November 2022).

**2.6. Model Accuracy Evaluation**

In order to evaluate the model accuracy, four statistical metrics were applied, namely the coefficient of determination ($R^2$), the Root Mean Square Error of calibration (RMSE), the Residual Prediction Deviation (RPD), and the Ratio of Performance to Inter-Quartile distance (RPIQ) as shown in the following equations (2)-(5). The $R^2$ measures the percentage of variance of the dependent variable as influenced by the independent variable. The RPD is explained as the ratio of standard deviation of the measured reference values to RMSE and it is used for NIR spectra in soil science as a value of correctness of the model. Also, RPIQ is explained as the ratio of the standard deviation of the inter-quartile distance of the measured reference data to RMSE.

A combination of the $R^2$ and RPD statistical metrics is allowing predictions to be favored or not. In more details, when $R^2 > 0.90$ and RPD > 3.0, excellent prediction is provided by the model. When $0.82 < R^2 < 0.90$ and $2.5 < RPD < 3.0$ a good approximation is feasible by the model. In addition, moderate approximation is made when $0.66 < R^2 < 0.82$ and $2 < RPD < 2.5$. Finally, poor distinction of high and low values are performed when R2 < 0.66 and RPD < 2 (Bellon-Maurel et al., 2010a; Saeys et al., 2005; Williams and Sobering, n.d.).

$$\boldsymbol{R^2}(y, \hat{y}) = 1 - \frac{\sum_{i=1}^{n}(y_i - \hat{y}_i)^2}{\sum_{i=1}^{n}(y_i - \bar{y})^2} \tag{3}$$

where $y_i$ and $\hat{y}_i$ are measured values and predicted values, respectively; n is the number of samples in the training set,

$$\text{RMSE} = \sqrt{\frac{1}{N}\sum(y_{pred} - y_{obs})^2} \qquad (4)$$

where N is the sample size, $y_{pred}$ is the predicted value, and $y_{obs}$ is the observed value. Typically, the model with the lowest RMSE is chosen.

$$\text{RPD} = \sqrt{\frac{\text{STDEV}(y_{obs})}{\text{RMSE}}} \qquad (5)$$

RPIQ is the ratio of performance to inter-quartile distance of the reference data in the external validation set). RPIQ is proposed to be applied instead of RPD in soil samples sets, for which often show a skewed distribution. As a result, RPIQ is a better way to standardize the RMSE in terms of population spread compared to RPD. RPIQ is based on quartiles representing the Q1 as the value below which 25% of the samples is found, Q2 as the value below where 50% of the samples are found and Q3 as the value below where 75% of the samples are found. Such an approach as described in equation 5 is useful to determine equivalent ranges of population spread (Bellon-Maurel et al., 2010b).

$$\text{RPIQ} = \sqrt{\frac{\text{IQ}}{\text{RMSE}}} \qquad (6)$$

where IQ = (Q3-Q1)

## 2.7 XRD Quantification

The samples were measured in a Bruker D8 Focus diffractometer with Bragg– Brentano configuration, operating at a voltage of 40 kV and an intensity of 40 mA. It contains a primary monochromator, working with a copper anticathode, Cu K$\alpha_1$ monochromatic radiation ($\lambda$ = 1.54056 Å). Acquisition was performed in 2θ scanning mode covering the 5-90° range with steps/sec of 0.02°. Samples were grinded until the fine powder was less than 20 μm. The powder was mounted in a circular quartz holder. The phases included in the refinement were all the minerals that were identified by Profex and BGNM libraries according to their peak intensities.

The Rietveld method (Rietveld, 1969) is a method that theoretically adjusts the structural and experimental parameters to the complete powder diffractogram profile of the sample, considering it as the sum of the Bragg reflections that appear at respective angular positions form crystalline material in the sample. It is considered as a total approach to mineralogical quantification. In addition, $\chi^2 < 5$ was considered as acceptable refinement statistics based on the complexity of soil samples examined in this study and based on previous similar work of clay material XRD analysis (Cuevas et al., 2022).

## 3. Algorithms and implementation

### 3.1 Partial Least Squares Regression

PLSR is a linear chemometric technique used for analysis of spectroscopic data for different applications. In our study, PLSR is used for the determination of soil carbonate content and is presented as a common modelling technique for quantitative spectroscopic analysis in soil mapping and classification as found in the literature (Kooistra et al., 2001; Martens and Næs, 1989; Viscarra Rossel, 2008). Decomposition of the spectral data into features (namely collective variables) is performed. The collective variables include most of the variance that exists in the reflectance NIR spectral data and thus linear models of the most correlated features are created.

$$X = T . P^T + Residuals(E) \qquad (7)$$

$$Y = T . C + Error(f) \qquad (8)$$

In PLSR, a decomposition of the *X* and *Y* variables with finding new latent variables and a selection of orthogonal factors that is maximizing the relation between prediction variables (X-soil reflectance) and response variables (Y-laboratory measured data) is performed. Components T that allows the decomposition of the predictors are searched by the PLSR (eq.1) and prediction of the response variables is also performed (equation 2). ***P*** and ***C*** are the factor loadings, and ***E*** and ***f*** are the residuals and errors matrices, respectively (Viscarra Rossel et al., 2006).

In our study, the PLSR was performed with the optimum 29 collective variables (CV's) in the existing NIR reflectance spectra acquired by the KSSL-USDA combined to LUCAS-TopSoil libraries.

**3.2 Cubist**

The Cubist chemometric technique is based on the M5 algorithm of Quinlan (Quinlan, 1992) and is widely and successfully applied in vis-NIR spectroscopy analysis. Thus, the Cubist method is considered as a competitive method to other methods of multivariate regression in terms of prediction accuracy.

The Cubist model is based on a regression tree construction with intermediate linear models extracted at each step of the procedure. It splits the original dataset that has similar attributes into subsets of sample and then generates multi-linear regression rules by optimal predictor variables selected from all spectral variables.

In our study, Cubist model was used with 29 collective variables (CV's), thus the optimum variables as described in PLSR method.

3.3 **Support Vector Machines**

Support vector machines (SVM) is a method that incorporates in its algorithm, linear equations for the regression analysis of multivariate cases (de Santana et al., 2021; Suykens et al., 2001).

The input parameters used for training the SVM are the NIR features that will be derived from the CV's calculated from the PLS regression model (CV's = 29). A linear kernel was used, and all spectra were normalized using a standard scaler (all values truncated between 0-1).

### 3.4 Multi-Layer Perceptrons

Multi-layer perceptrons (MLPs) or deep feedforward networks are machine learning models that use intermediate computations to transform their input $x$ to the output $y$ and in doing so evaluate a function $f$:

$$y = f(x; \boldsymbol{\theta}) \tag{9}$$

The transformations (linear and non-linear) that each model's layer implements to the data is parameterized by its weights $\boldsymbol{\theta}$. In this context, the model's goal is to find the set of values for the weights $\boldsymbol{\theta}$ of all layers in the network that correctly map inputs instances to their corresponding targets. Put in another way, the weights $\boldsymbol{\theta}$ that minimize the difference between the distribution of the output $y$ and the true underlying distribution of the targets.

In contrast to linear models and to extend them to represent nonlinear functions of $x$, MLPs apply the linear model to a transformed input $\phi(x)$ and not to $x$. The ultimate goal is to learn $\phi$:

$$y = f(x; \boldsymbol{\theta}; w) = \phi(x; \boldsymbol{\theta})^{\top} w \tag{10}$$

where $\boldsymbol{\theta}$ are parameters used to learn $\phi$ from a broad family of functions and parameters $w$ that map from $\phi(x)$ to the desired output. The choice of how to represent the output determines the form of the cost function which usually expresses the difference between the predicted and the true distribution. In most cases the parametric model defines a distribution $p(y \mid x; \boldsymbol{\theta})$ and the principle of maximumm likelihood is used:

$$J(\theta) = -E_{x,y \sim \widehat{p_{data}}} \log p_{model}(y|x) \tag{11}$$

In such cases, the cost function is the negative log-likelihood between targets and model's predictions. A simpler approach is to merely predict a statistic of y conditioned on x. Note that the cost function will often combine a regularization term such as weight decay or dropout layers in order to avoid overfitting.

Our MLP model consists of three layers of 500, 200 and 50 units respectively (Figure 1). Since we are predicting only one soil attribute ($CaCO_3$) the model has a single output. To introduce non-linearity we apply the Rectified Linear Unit (ReLU) activation function and as a measure to mitigate overfitting $L_1$ and $L_2$ regularizers are applied to the third layer. We use the Adam optimizer (Kingma and Ba, 2014)with a decaying learning rate. Savitzky-Golay second derivative smoothing filter is applied to the spectral dataset before being fed to the network.

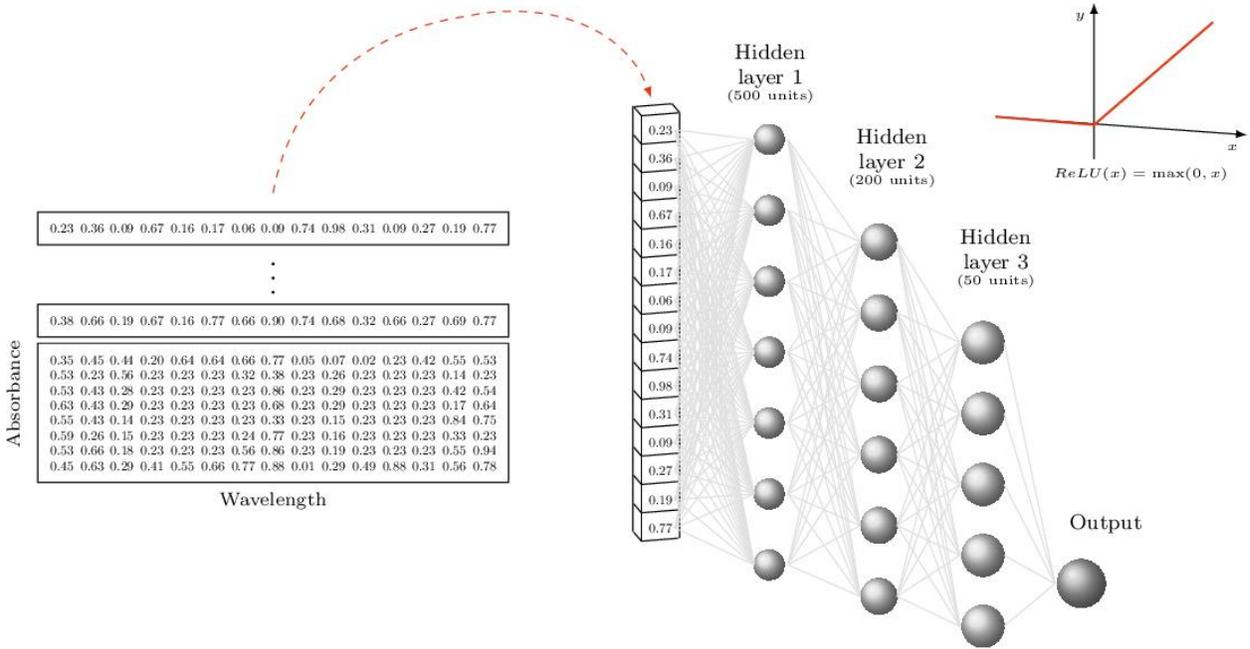

*Figure 1. MLP architecture. Three consecutive hidden layers (with 500, 200, 50 units) are filtering the combined spectral dataset in the NIR region, to a final output. An activation function ReLU is applied to facilitate training of the MLP model.*

### 3.5	Convolutional Neural Networks

Convolutional neural networks (CNN) are neural networks specialized in grid-like topology data i.e. images that are represented as 2D grid of pixels. Their architecture consists of stacked convolutional layers of multiple filters that convolve with their input to produce a series of feature maps. Generally, a convolution is an operation on two functions:

$$s(t) = \int (x * w)(t) \tag{12}$$

where $x$ is the input function and $w$, also known as the kernel (or filter), needs to be a valid density function for the output to be a valid weighted average of $x$ over time $t$.

When an instance passes through a CNN layer it usually undergoes three stages: first, the layer applies several convolutions in parallel to produce a set of linear activations. Then, each linear activation passes through a nonlinear activation function, such as a rectified linear unit. Finally, the output is further modified by a pooling function which serves both as a downsampling technique to reduce statistical and computational burden (Goodfellow et al., 2016) as means to make the representation become approximately invariant to small translations of the input (Goodfellow et al., 2016). Besides being translation invariant, CNNs learn spatial hierarchies of patterns: lower layers learn smaller patterns whereas layers near the output learn more complex patterns and abstract visual concepts (Chollet, 2021).

These traits make CNNs an excellent choice for machine vision tasks such as object detection, image classification and semantic segmentation. Although designed for multidimensional data e.g. images and CT scans, CNNs perform equally good on one-dimensional inputs. In fact, CNNs have been used with relative success in tasks that involve sequences e.g. time-series. However, in our case we applied CNNs to the sample spectrograms (2D images) although we trained a CNN on spectra which yielded results on par with the MLP Regressor.

Our CNN model is made of three convolutional layers of 32, 64 and 128 filters respectively, each succeeded by a max-pooling layer (F). All convolutional and max-pooling kernels are 3X3 size. A dense layer of 50 units sits on top of the convolutional stack which in turn is connected to the network's single output. As in the MLP approach, we use ReLU activation function and Adam optimizer. In this approach, the network's input is not spectra but their respective spectrograms, each with dimensions 244X488 pixels.

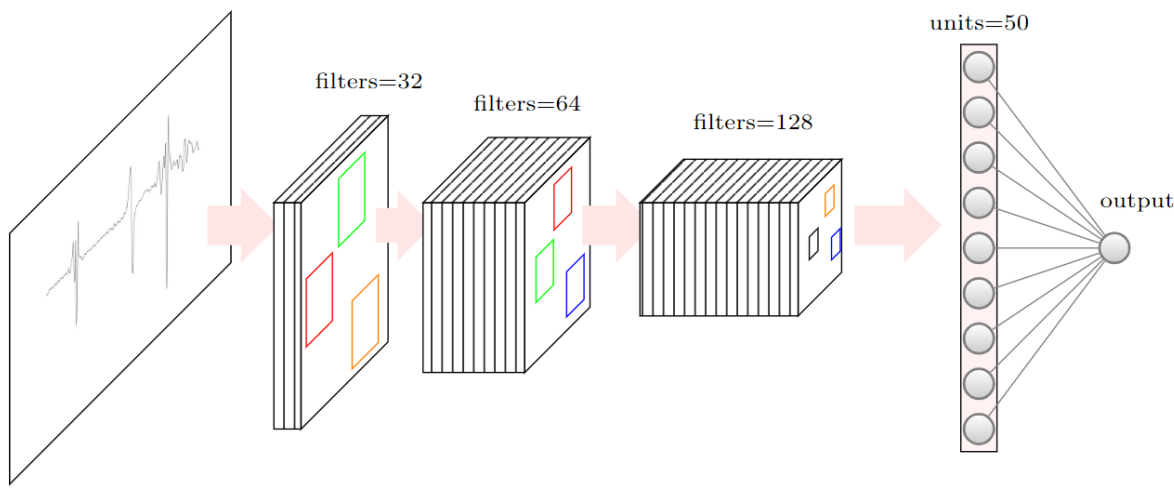

*Figure 2. CNN architecture. Three consecutive convolutional layers (with 32, 64, 128 units) are filtering the combined spectral dataset as images in the NIR region, with dimensions 244x488 pixels, to a final output. An activation function ReLU is also applied to facilitate training of the CNN model.*

**4.1 Results and Discussion**

Calibration models using two large spectral databases combined, that are obtained by the USDA (KSSL spectral library) and the European Union (LUCAS TopSoil spectral library) as input for the commonly used spectral regression methods (PLSR, Cubist, SVM) and, for the first time, for the machine learning algorithms namely MLP and CNN, and furthermore metrics models on the prediction set are achieved with high performance statistics ($R^2$ = 0.84, RPD = 2.14 for MLP) in the prediction set and with $R^2$ = 0.68 and RPD = 1.47 for CNN, for the prediction accuracy of soil carbonates, in the second derivative of NIR spectra.

Soil carbonates mainly calcium and magnesium carbonates but also their hydrocarbonates have several absorption peaks in the NIR region, which are due to overtone and combination bands of the $CO_3^{2-}$ fundamental bands, that occur in mid-IR (Clark et al., 1990). In the NIR, carbonates possess five very characteristic bands. Firstly, a prominent peak occurs near 2340 nm that is an overtone of asymmetrical stretching $v_3$, 1415 cm$^{-1}$ in the mid-IR region. Furthermore, the peak near 2500-2550 nm is related to a combination of symmetrical stretching ($v_1$) and the first overtone of asymmetrical stretching ($v_3$), exactly as ($v_1 + 2v_3$). Some weaker absorptions occur near 1415 nm that occurs due to the crystallized water, around 1900 nm ($v_1 + 3v_3$), near 2000 nm ($2v_1 + 2v_3$) and near 2160 nm ($3v_1 + 2v_4$), where $v_4$ is the in-plane bending ($v_4$, 680 cm$^{-1}$) (Hunt, 1977). It is important to note that the position of these absorption bands varies with composition (Hunt and Salisbury, 1970). Few studies quantify their composition in soil except the studies of Ben-Dor (1990), Khayamim (2015) and Barthes (2016) who used vis–NIR to estimate the carbonate content in soils (Barthès et al., 2016; Ben-Dor and Banin, 1990; Khayamim et al., 2015). A recent study examined the soil carbonates content using mid-IR spectroscopy in addition to utilizing KSSL library as calibration dataset (Comstock et al., 2019).

Soil samples (no.45) of both Xatzisavva (SAM-1) wine fields in Alexandroupolis in Greece and Vourvoukelis (SAM-2) in Avdira in Greece were collected and samples (no.19) from SAM-1 and SAM-2 groups exhibited non-zero values in carbonate content, using the volumetric method as described in the methods section and were recorded (g/100g). As shown in Table 2, at least one soil sample from SAM-1 group is rich in carbonates (highest content = 18.14%) and the poorest carbonates content is in SAM-2 group (lowest content = 0.04%).

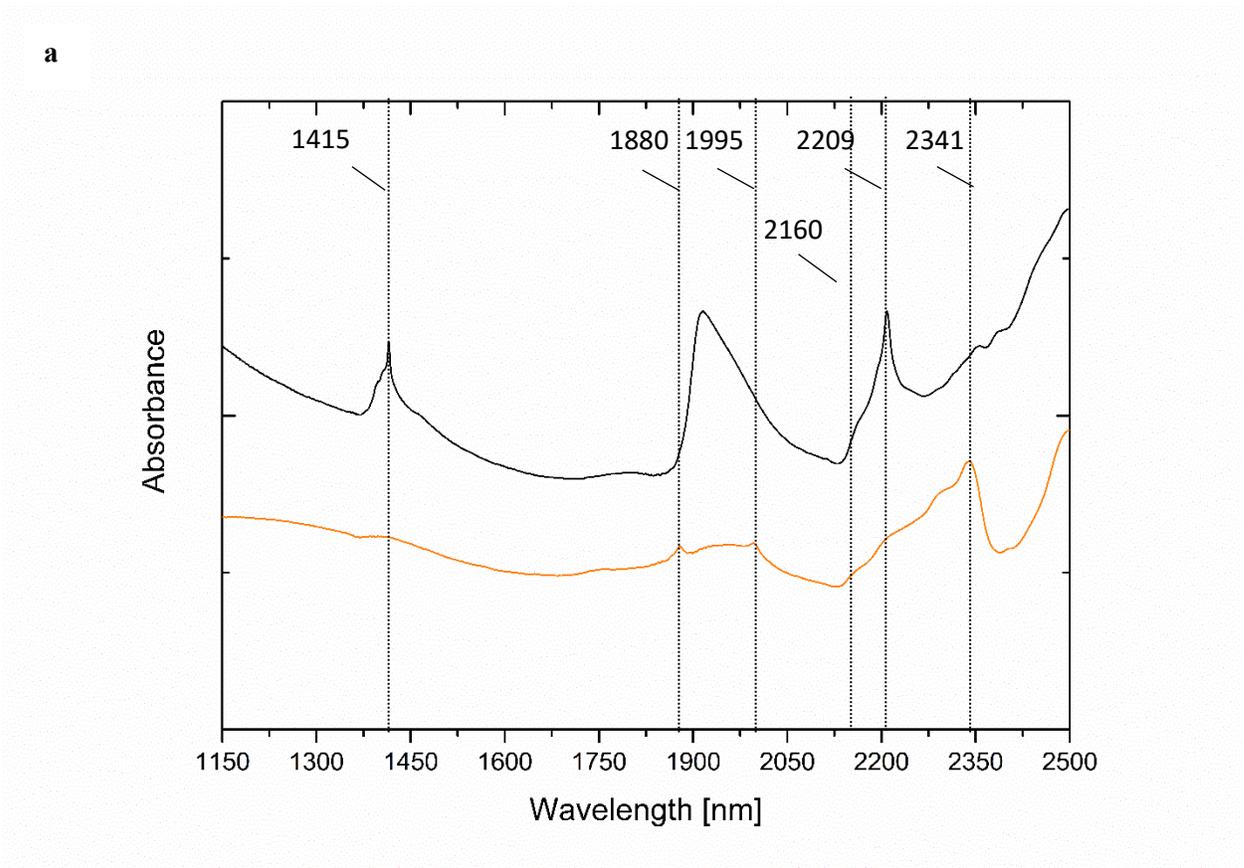

**b**

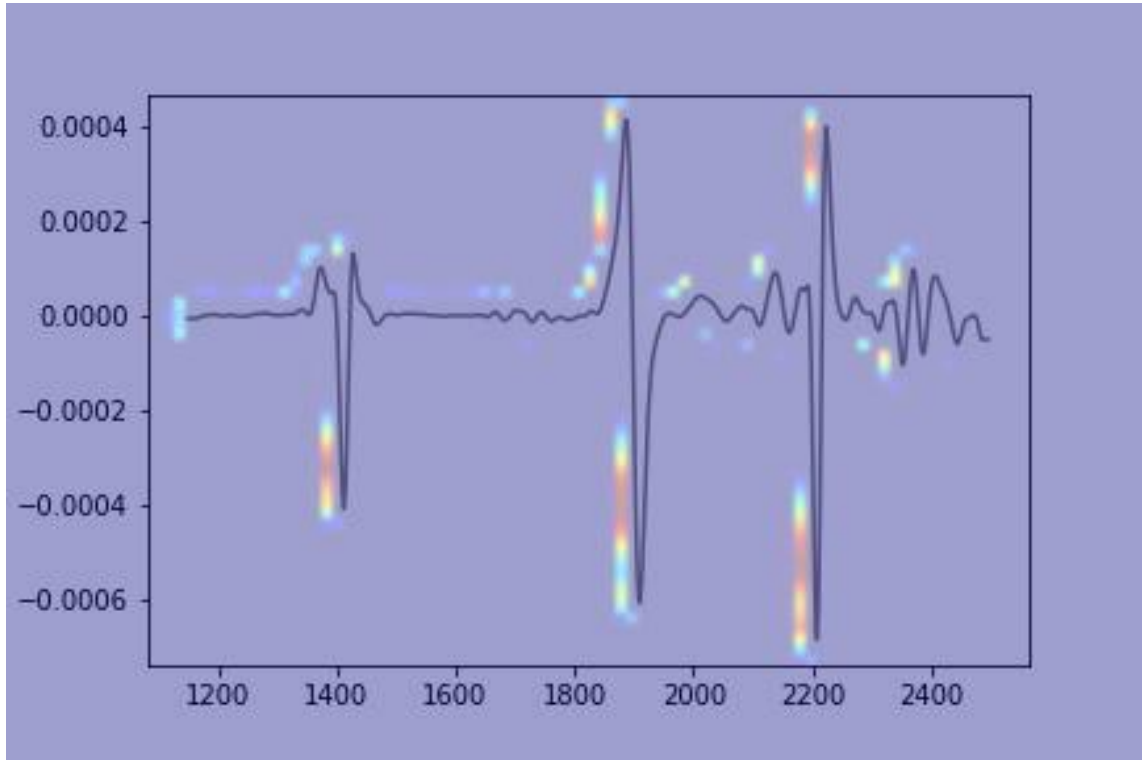

*Figure 3: On the top (a) absorbance spectra of an indicative soil sample (s03) (black) of Asyrtiko species from the Xatzisavva wine field in Alexandroupolis and calcium carbonate (Sigma, 239216) white powder spectrum (orange). Indication lines of carbonates vibration peaks are also shown as dashed lines. (b) a saliency map from CNN model in second derivative spectra, is showing the most favorable peaks that are considered to trigger the CNN model in a color gradient from red (most favored peaks) to light blue (less favored peaks).*

| S/no | Exp g/100g | Pred g/100g | |
|------|------|------|------|
| S01 | 7.85 | 11.53 | |
| S02 | 1.83 | 0.86 | |
| S03 | 8.56 | 8.80 | |
| S04 | 10.84 | 5.85 | |
| S05 | 9.32 | 8.37 | |
| S06 | 4.63 | 1.64 | |
| S07 | 3.39 | 1.52 | |
| S08 | 5.75 | 3.31 | SAM-1 |
| S09 | 1.18 | 1.07 | |
| S10 | 18.14 | 13.44 | |
| S11 | 11.28 | 14.18 | |
| S12 | 17.32 | 14.38 | |
| S13 | 17.00 | 18.57 | |
| S14 | 1.12 | 1.80 | |
| S15 | 0.07 | 0.60 | |
| S16 | 0.13 | 0.59 | |
| S17 | 0.13 | 0.53 | SAM-2 |
| S18 | 0.09 | 0.51 | |
| S19 | 0.04 | 0.51 | |

*Table 1:* *Experimental values of diverse soil samples (no.19) using the volumetric method in comparison with the predicted carbonates values from MLP model. Two groups of soil samples (SAM-1 and SAM-2) were measured and predicted.*

|  | **PLSR** | **SVM** | **Cubist** | **MLP** | **CNN (images)** |
|---|---|---|---|---|---|
| Second Derivative | $R^2 = -1.43$ | $R^2 = -0.19$ | $R^2 = 0.45$ | $R^2 = 0.84$ | $R^2 = 0.68$ |
|  | RMSE = 9.42 | RMSE = 6.59 | RMSE = 4.47 | RMSE = 2.11 | RMSE = 4.11 |
|  | RPD = 0.64 | RPD = 0.91 | RPD = 1.35 | RPD = 2.14 | RPD = 1.47 |
|  | RPIQ = 1.00 | RPIQ = 1.43 | RPIQ = 2.11 | RPIQ = 3.33 | RPIQ = 2.29 |

*Table 2:* *Regression prediction models applied to all data for carbonates (two spectral libraries, firstly 6,833 NIR spectra from KSSL (USA) combined with 21,782 NIR spectra from Lucas TopSoil (all European fields) for carbonates). Statistics of $R^2$, Root Mean Square Error (RMSE), Residual Prediction Deviation (RPD) and Ratio of Performance to Inter-Quartile distance (RPIQ) are shown.*

In an attempt to determine which wavelength absorbance peaks, possess more weight on the model's performance and due to the fact that it is easier to visualize this through saliency mappings (peaks on the second derivative absorbance spectra) we develop a CNN model as discussed previously. The most favored peaks are in the 1415 nm, in the 1908 nm, in the 2209 nm and in the 2335 nm. The sharp band at 1415 nm that is shown in soil NIR spectra in Figure 3, is due to calcite and hydromagnesite ($Mg_5(CO_3)_4(OH)_2 \cdot 4(H_2O)$) that is a mineral also present in calcareous soils especially in SAM-1 group and absent in SAM-2 group. Hydrous carbonates, like other hydrous minerals, typically exhibit strong bands around 1400 nm due to O–H stretching or a combination of the symmetric H–O–H stretch and H–O–H bend (Harner and Gilmore, 2015). The shape of the absorption peaks can indicate whether this feature is produced by a hydroxyl or water group. The characteristic band near 1415 nm is also due to kaolinite that shows absorption wavelengths near 1400 nm (1395 and 1415 nm) that are overtone vibrations of the O–H stretch near 2778 nm (3600 cm$^{-1}$), and can to one part be attributed to the first overtone of structural O–H stretching mode in its octahedral layer (Stenberg et al., 2010). The 1880 nm band is due to $v_1 + 2v_3$, where $v_1$ is the totally symmetric C-O stretch and $v_3$ is the doubly degenerate antisymmetric C-O stretching mode occurring near 7000 nm (Hunt, 1977).

Also, at 1908 nm occurs a combination of H–O–H bending and the asymmetrical stretching fundamentals. This band is possible overlapping with the main band of other minerals and in particular O-H stretch of kaolinite. Also, the peak near 2209 nm is triggered by the presence of calcite. Lastly, around 2235 nm a prominent peak of calcite triggers the saliency map derived from CNN model. Other peaks that are triggered by the content of carbonates by visual inspection of the saliency map, are difficult to assign because of various peak overlaps with hydroxyl and/or water molecules bound to mineral crystallites. As a result, NIR spectral assignment towards elucidation of carbonates peaks is a multiparametric task, especially in complex samples such as soil samples.

In order to describe in full detail, the carbonates content of the SAM-1 (sample S03), and SAM-2 (S18) we took insight in the quantification of the minerals by means of powder X-ray diffraction (XRD) method. Diffractograms were acquired in an in-house diffractometer, as mentioned in the methods section. Rietveld refinement using structural models was then performed.

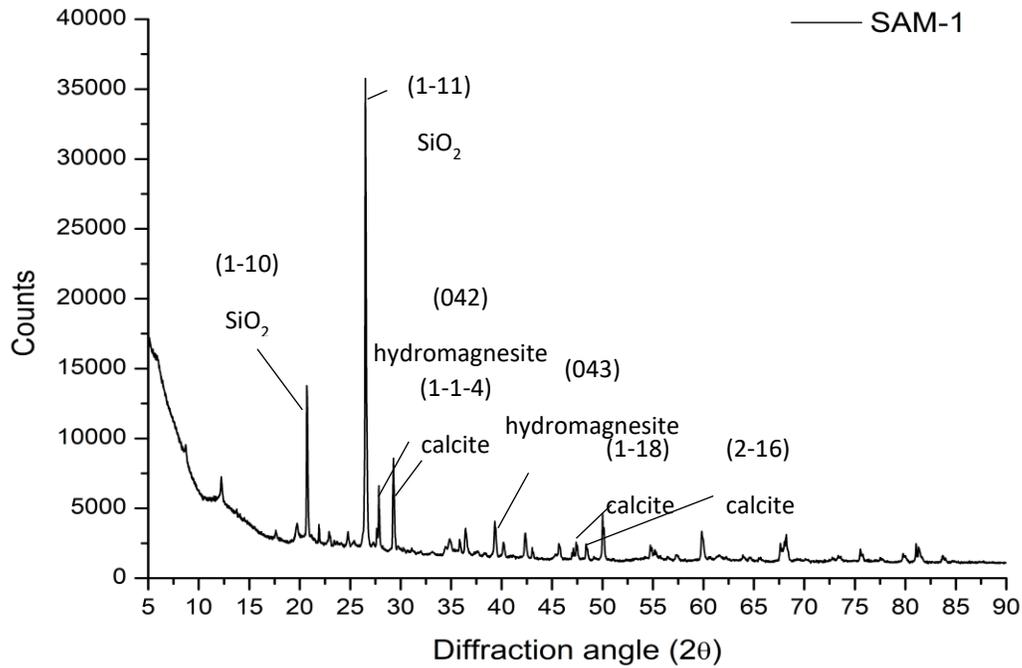

*Figure 4: XRD diffractogram shows that peaks of characteristic planes of calcite, and hydromagnesite are present in SAM-1, among other distinct minerals peaks. Crystalline Index (CI) is calculated to 0.72.*

Quantification of carbonates minerals in crystalline phase, in the representative soil sample (S03) of SAM-1 group, shows that total carbonates are contained in 6.07 % (w/w) with a crystalline index = 0.72. The total carbonates content is then calculated to 8.43 % (w/w). The minerals found were firstly calcite 3.88 % (w/w), and also hydromagnesite 2.19 %, as shown in Table 3. A total of 8.43 % (w/w) in ratio of total minerals and in specific of both their crystalline phases of SAM-1 group and amorphous content based on crystalline index, is in good agreement with both volumetric method and MLP prediction of NIR spectra based on the calibration libraries of KSSL (USA) and LUCAS-Topsoil (EU).

| SAM-1 | Molecular formula | Phase quantification from XRD (wt-%) | Volumetric method (wt-%) | MLP (wt-%) |
|---|---|---|---|---|
| **Calcite** | $CaCO_3$ | 3.88 | - | - |
| **Hydro-magnesite** | $Mg_5(CO_3)_4(OH)_2 \cdot 4H_2O$ | 2.19 | - | - |
| **Total carbonates** | $CO_3^{2-}$ | 6.07 (Crystalline Index = 0.72) Total = 8.43 | **8.56** | **8.80** |

*Table 3: Comparison of total carbonates content of S03 from SAM-1 group by volumetric method, MLP prediction and XRD phase quantification of both calcite and hydromagnesite minerals shows a very good values agreement.*

In SAM-2 group, the carbonates content is negligible, as shown in Figure 5 and the XRD diffractogram where all peaks of calcite are absent, at 29.4º for (1-1-4) calcite plane and around 47.4º and 48.5º for (2-2-2) and (1-18) planes, respectively. Also, hydromagnesite is absent at 27.7º for (042) and at 39.2º for (080).

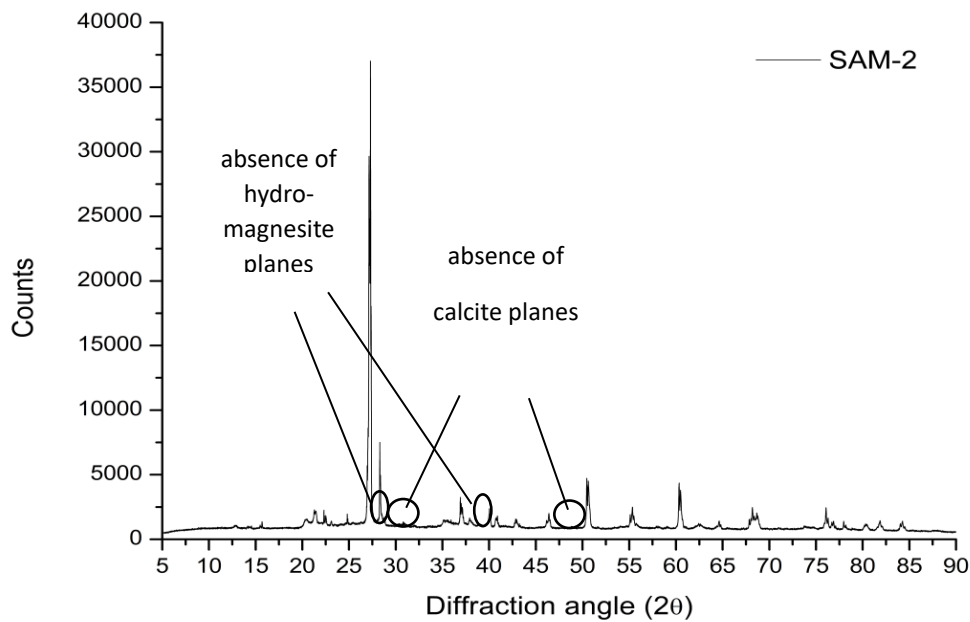

*Figure 5: Absence of calcite and hydromagnesite crystalline phases in diffractogram for SAM-2 group representative sample (S18).*

4. **Conclusions**

Carbonates content of soil is an essential soil chemical property, with a significant impact on growth and production of crops. Soil genesis records and soil classifications as shown in various soil profiles are also mostly affected by the distribution of carbonates content of soils. Here we present a modified way of soil carbonates prediction other than classical volumetric method. A minimal input of a NIR spectra provides the total carbonates content in soil samples, as calibrated with two spectral libraries, namely USDA-NRCS-KSSL and EU-LUCAS-TopSoil with 28,615 NIR spectra, where MLP method is trained in this dataset and applied for prediction. MLP method performed good in a diverse test set ranging from diverse carbonates content as low as 0.07%, to high content (18.14%) in soil samples. CNN method produced less favorable results in terms of prediction with experimentally derived carbonates values and a saliency map showed us the peaks triggered in training and prediction spectral sets. Peak assignment of various carbonates and hydrous carbonates minerals in the NIR region was also established. The presence of hydromagnesite in soil samples with highest carbonates content (SAM-1) was verified by means of XRD Rietveld analysis. Quantification of the total carbonates content in SAM-1 group with XRD, and in representative sample S03, is in good agreement with MLP prediction and volumetric method applied, while the absence of carbonates content is obvious in SAM-2 group with XRD analysis.

## 5. Acknowledgments

This study was supported by the project "AGRO4+" - Holistic approach to Agriculture 4.0 for new farmers" (MIS 5046239) which is implemented under the Action "Reinforcement of the Research and Innovation Infrastructure", funded by the Operational Programme "Competitiveness, Entrepreneurship and Innovation" (NSRF 2014-2020) and co-financed by Greece and the European Union (European Regional Development Fund).

Authors would like to thank Dr. Nikolaos Kazakis, Dr. Nestor Tsirliganis and Dr. Chairi Kiourt for access to instrumentation and the staff at the USDA-NRCS-KSSL and EU-LUCAS-TopSoil for the data provided in this study.

**Code availability section**

Carbonates_prediction

Contact: petros.tamvakis@athenarc.gr, +1-262-960-2449 (USA)

Hardware requirements: CPU/GPU

Program language: Python 3

Software required: https://www.tensorflow.org/ and dependencies therein.

Program size: 110 KB

The source codes are available for downloading at the link: https://github.com/petamva/carbonates_prediction

List of Figures

1. *Figure 2. MLP architecture. Three consecutive hidden layers (with 500, 200, 50 units) are filtering the combined spectral dataset in the NIR region, to a final output. An activation function ReLU is applied to facilitate training of the MLP model.*

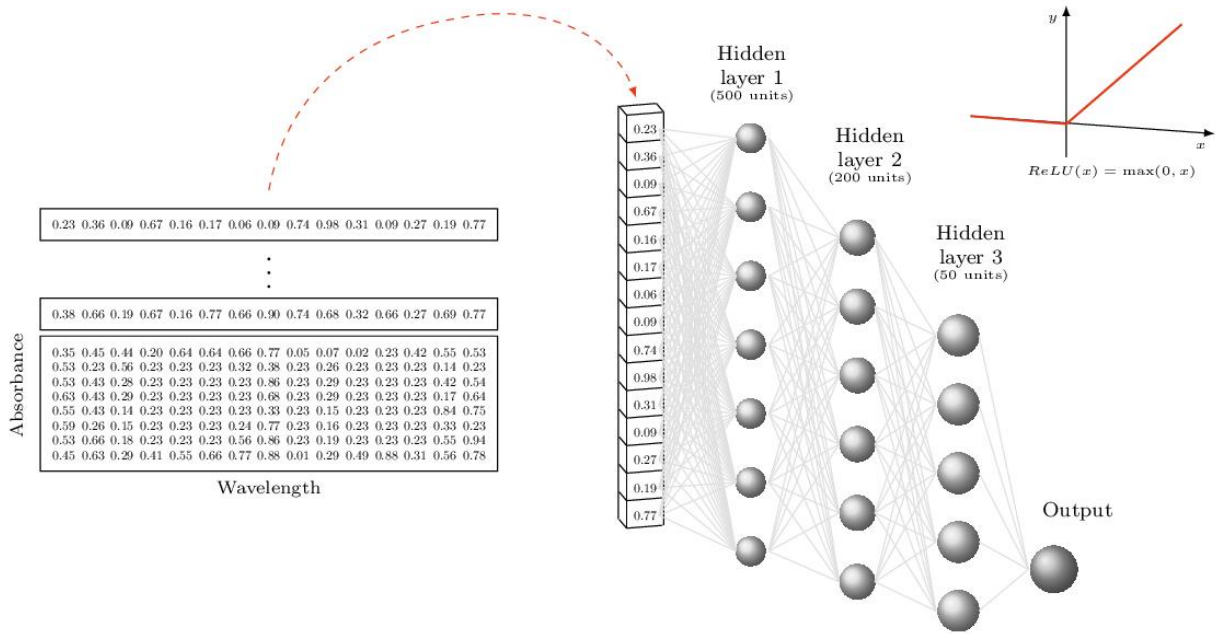

2. *Figure 2. CNN architecture. Three consecutive convolutional layers (with 32, 64, 128 units) are filtering the combined spectral dataset as images in the NIR region, with dimensions 244x488 pixels, to a final output. An activation function ReLU is also applied to facilitate training of the CNN model.*

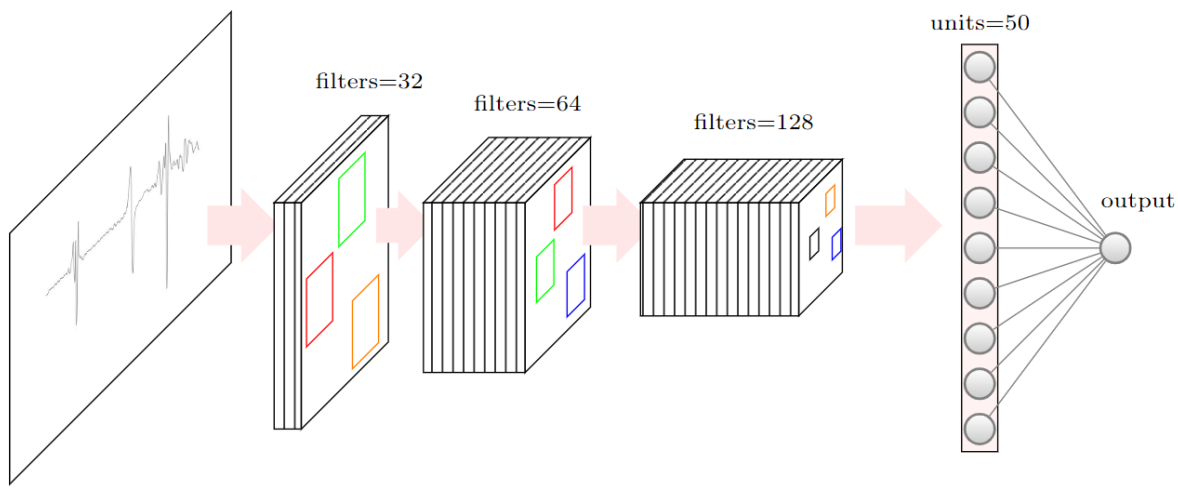

3. *Figure 3: On the top (a) absorbance spectra of an indicative soil sample (s03) (black) of Asyrtiko species from the Xatzisavva wine field in Alexandroupolis and calcium carbonate (Sigma, 239216) white powder spectrum (orange). Indication lines of carbonates vibration peaks are also shown as dashed lines. (b) a saliency map from CNN model in second derivative spectra, is showing the most favorable peaks that are considered to trigger the CNN model in a color gradient from red (most favored peaks) to light blue (less favored peaks).*

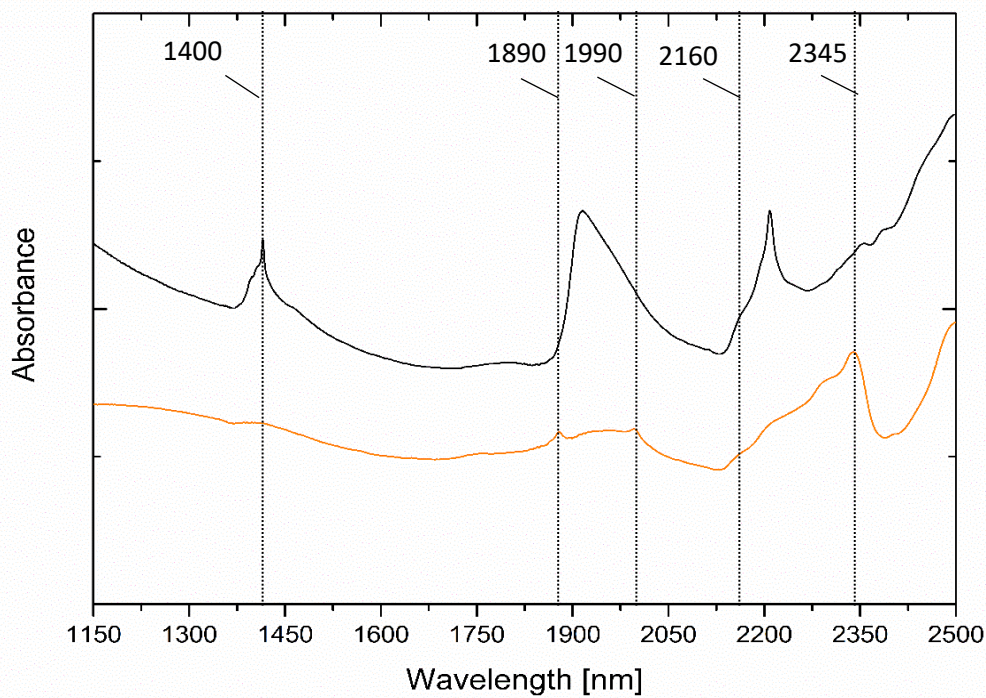

a

b

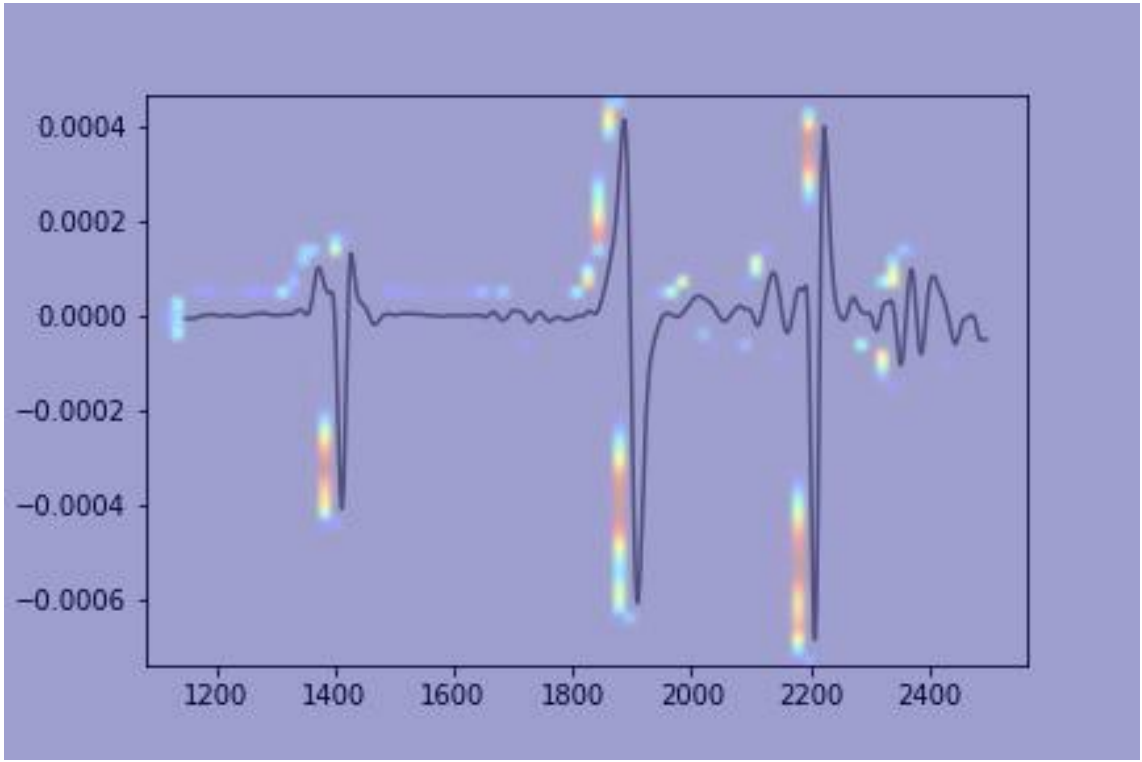

4. *Table 1: Experimental values of diverse soil samples (no.19) using the volumetric method in comparison with the predicted carbonates values from MLP model. Two groups of soil samples (SAM-1 and SAM-2) were measured and predicted.*

| S/no | Exp g/100g | Pred g/100g | |
|---|---|---|---|
| S01 | 7.85 | 11.53 | |
| S02 | 1.83 | 0.86 | |
| S03 | 8.56 | 8.80 | |
| S04 | 10.84 | 5.85 | |
| S05 | 9.32 | 8.37 | |
| S06 | 4.63 | 1.64 | |
| S07 | 3.39 | 1.52 | |
| S08 | 5.75 | 3.31 | SAM-1 |
| S09 | 1.18 | 1.07 | |
| S10 | 18.14 | 13.44 | |
| S11 | 11.28 | 14.18 | |
| S12 | 17.32 | 14.38 | |
| S13 | 17.00 | 18.57 | |
| S14 | 1.12 | 1.80 | |
| S15 | 0.07 | 0.60 | |
| S16 | 0.13 | 0.59 | |
| S17 | 0.13 | 0.53 | |
| S18 | 0.09 | 0.51 | SAM-2 |
| S19 | 0.04 | 0.51 | |

5. *Table 2: Regression prediction models applied to all data for carbonates (two spectral libraries, firstly 6,833 NIR spectra from KSSL (USA) combined with 21,782 NIR spectra from Lucas TopSoil (all European fields) for carbonates). Statistics of $R^2$, Root Mean Square Error (RMSE), Residual Prediction Deviation (RPD) and Ratio of Performance to Inter-Quartile distance (RPIQ) are shown.*

|  | **PLSR** | **SVM** | **Cubist** | **MLP** | **CNN (images)** |
|---|---|---|---|---|---|
| Second Derivative | $R^2 = -1.43$  RMSE = 9.42  RPD = 0.64  RPIQ = 1.00 | $R^2 = -0.19$  RMSE = 6.59  RPD = 0.91  RPIQ = 1.43 | $R^2 = 0.45$  RMSE = 4.47  RPD = 1.35  RPIQ = 2.11 | $R^2 = 0.84$  RMSE = 2.11  RPD = 2.14  RPIQ = 3.33 | $R^2 = 0.68$  RMSE = 4.11  RPD = 1.47  RPIQ = 2.29 |

6. *Figure 4: XRD diffractogram shows that peaks of characteristic planes of calcite and hydromagnesite are present in SAM-1, among other distinct minerals peaks.*

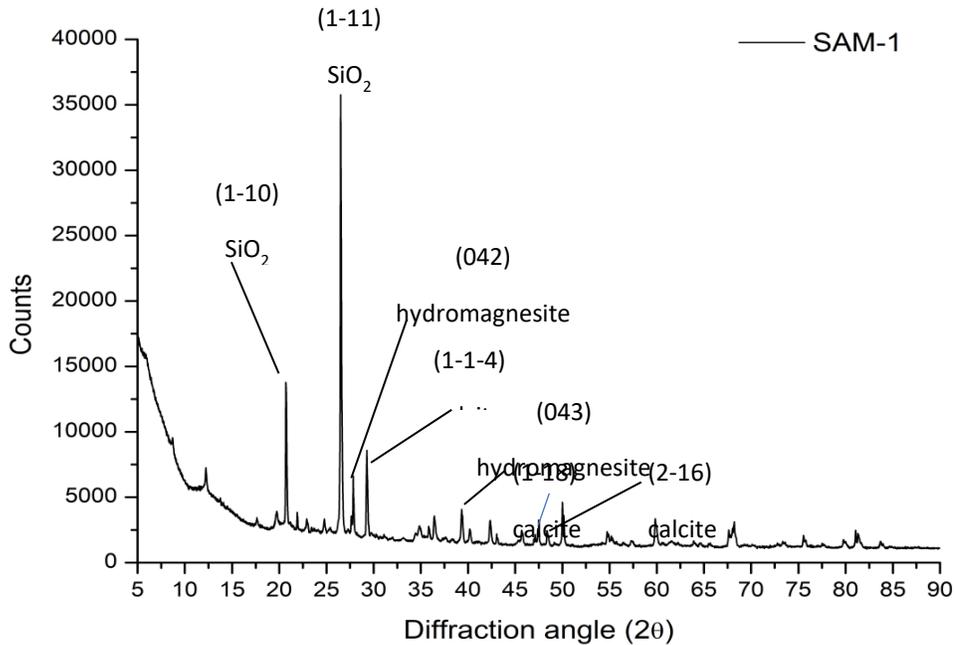

7. ***Figure 5:*** *Absence of calcite and hydromagnesite crystalline phases in diffractogram for SAM-2 group representative sample (S18).*

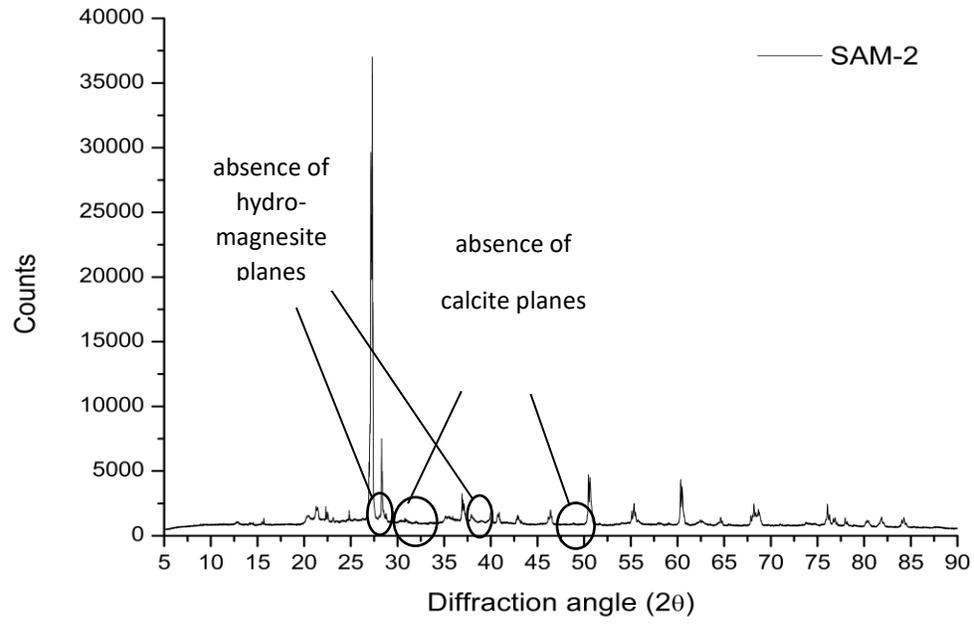

8. *Table 3: Comparison of total carbonates content of S03 from SAM-1 group by volumetric method, MLP prediction and XRD phase quantification of both calcite, and hydromagnesite minerals shows a very good values agreement.*

| SAM-1 | Molecular formula | Phase quantification from XRD (wt-%) | Volumetric method (wt-%) | MLP (wt-%) |
|---|---|---|---|---|
| **Calcite** | $CaCO_3$ | 3.88 | - | - |
| **Hydro-magnesite** | $Mg_5(CO_3)_4(OH)_2 \cdot 4H_2O$ | 2.19 | - | - |
| **Total carbonates** | $CO_3^{2-}$ | 6.07 (Crystalline Index = 0.72) Total = 8.43 | **8.56** | **8.80** |